\documentclass{article}
\usepackage{spconf,amsmath,graphicx,subfigure}
\usepackage{booktabs}

\usepackage{multirow}
\usepackage{array}
\usepackage{xcolor}
\usepackage{chngpage}
\usepackage[symbol]{footmisc}
\usepackage{booktabs}
\usepackage[ruled,vlined]{algorithm2e}

\usepackage{cite}
\usepackage{algorithmic}
\usepackage{graphicx}
\usepackage{textcomp}
\title{Emotion Recognition with Incomplete Labels Using Modified Multi-task Learning Technique
}

\name{Phan Tran Dac Thinh$^{1}$, Hoang Manh Hung$^{1}$\thanks{1 Contributed equally},  
 Hyung-Jeong Yang, Soo-Hyung Kim, and Guee-Sang Lee$^{*}$\thanks{* Corresponding author}
}
\address{Department of Artificial Intelligence Convergence, Chonnam National University, South Korea\\ 
}
\begin{document}
\maketitle
\begin{abstract}
The task of predicting affective information in the wild such as seven basic emotions or action units from human faces has gradually become more interesting due to the accessibility and availability of massive annotated datasets. In this study, we propose a method that utilizes the association between seven basic emotions and twelve action units from the AffWild2 dataset. The method based on the architecture of ResNet50 involves the multi-task learning technique for the incomplete labels of the two tasks. By combining the knowledge for two correlated tasks, both performances are improved by a large margin compared to those with the model employing only one kind of label.
\end{abstract}

\section{INTRODUCTION}
Affective computing aims to transfer the understanding of human feelings to computers, so they could recognize humans’ emotional states and be applied to multiple advanced areas such as education or health service. Affective states can be decided by a wide range of sources in three main categories, namely visual, auditory and biological signals. Visual information, especially facial clues, is the most important and most adopted data due to high availability, interpretability and strong pertinence to emotional states. \\
To analyze the affective states, Ekman \cite{ekman1971constants} introduced the six basic emotions, i.e., anger, disgust, fear, happiness, sadness and surprise. Those categorical values are extensively applicable to human beings but this is not the only way to perceive the emotional states. In terms of dimensional model, they can be represented as continuous values, namely valence and arousal. Valence shows how positive or negative the emotion is while arousal measures the agitation level which is from non-active to ready to act. Furthermore, according to the Facial Action Coding System (FACS) \cite{friesen1978facial}, facial movements which are defined as Action Units (AUs) are recorded to interpret emotions.\\ 
Affective Behavior Analysis In-The-Wild (ABAW) 2021 \cite{kollias2021analysing} is a competition with the primary goal of improving the machines’ capability of understanding human feelings, emotions and behaviors. The competition provides the massive dataset called AffWild2 \cite{kollias2020analysing, kollias2021distribution, kollias2021affect, kollias2019expression, kollias2019face, kollias2019deep, zafeiriou2017aff} about emotions in the wild with annotations for seven basic emotions, AUs and valence/arousal. There are three tasks corresponding to the three types of annotations and the dataset contains the videos and also the cropped and aligned images extracted from those videos.\\
In this paper, since the dataset does not have complete labels for seven basic expression classification and facial action unit detection, we propose a modified multi-task learning technique for ResNet50 as the main model for implementing both tasks. Moreover, the data for seven emotions is highly imbalanced towards more common emotions such as neutral and happy, so we employ the Focal Loss to counter this effect. The detail of our work will be described in the next section.

\section{Proposed Method}
\subsection{Preprocessing}
The AffWild2 dataset provides the cropped and aligned images that are extracted from the videos. We use them for both training and validation stages and did not use other tools to acquire the images from the videos. The input size for the model is 112 x 112 and RGB color space is applied. The images are normalized before inputting to the model and no augmentation techniques are used to enlarge the dataset. The audios are not adopted for training in our method since not all videos contain sounds and the sounds in some cases are noise from the environment or human activities. Without proper processing or an adequate mechanism to analyze the audios, they probably cause ambiguity and drop of performance to the main model. 

\subsection{Model Structure}
ResNet50 \cite{he2016deep} is the backbone of our deep learning network. Commonly, the pretrained weights on the ImageNet \cite{russakovsky2015imagenet} are used to accelerate and enhance the training performance. In the field of emotion recognition, thanks to the existing works on emotion recognition, we opt for the pretrained weights on the VGGFace2 dataset \cite{cao2018vggface2}. The VGGFace2 dataset is not only large in the amount of images but also varied in the number of subjects and covers a large range of pose, age and ethnicity too. This coincides with the concept of AffWild2 dataset which is not dependent on the context nor the age, gender, ethnicity, social status, etc. The fully connected layer is cut and then the pretrained weights are loaded on the main backbone. A new dense layer with 512 and two dense layers of 7 or 12 neurons are sequentially added according to the specific task. \\
The model takes the input as static images. Because a lot of images do not have both labels for seven emotions and AUs, we need to have a particular training scheme to better learn the shared knowledge between two tasks. The training scheme is shown as below:
\begin{algorithm}
    \SetAlgoLined
    % \KwInput{Write here the result }
    \textbf{Input}: \\
    \quad Set images which has 7 expression as target $E$ \\
    \quad Set images which has 12 action units as target $A$ \\
    \quad Set images which has both types of labels $B$ \\
    \quad Number of epochs $T$\\
    \textbf{Output}: \\
    \quad One label for 7 expression $e$ \\
    \quad Multi-labels for 12 action units $a$ \\
    
    \For{$t = 1, t \leq T$}{ 
        \For{$i = 1, i \leq len(E)$}{
            $e = Model(E_{i})$\\
            $Loss = F(e, \theta _{i})$\\
            Update weight $\theta _{i}$
        }
        \For{$i = 1, i \leq len(A)$}{
            $a = Model(A_{i})$\\
            $Loss = F(a, \theta _{i})$\\
            Update weight $\theta _{i}$
        }
        \For{$i = 1, i \leq len(B)$}{
            $e, a = Model(A_{i})$\\
            $Loss = F(e, a, \theta _{i})$\\
            Update weight $\theta _{i}$
        }
    }
    \caption{The training strategy for proposed method}
\end{algorithm} 

\subsection{Loss Function}

For seven basic emotion classification, we use Focal loss \cite{lin2017focal}. The dataset for expression classification is imbalanced so we apply this loss to deal with this problem. 
For facial action unit detection, we use binary cross entropy loss for each action unit. 
The total loss is the summation of the two above losses with the same weight. 

\begin{figure}[ht]
\begin{center}
\includegraphics[scale=0.6]{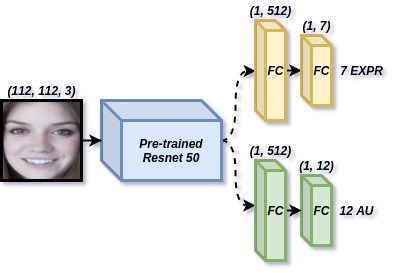}
\caption{Overview system of proposed method}
\label{fig:vgg}
\end{center}
\end{figure}

\section{Experimental Results}
\subsection{Training Setup}
The training process is optimized by the Adam optimizer. We use GPU RTX 2080Ti as the hardware and Pytorch framework as the software. The mini-batch has size 256. To regularize the training process and accelerate the convergence of the model, we use the Cosine Annealing as the learning rate scheduler with the starting learning rate of 0.001. The testing models for the most time achieve the best results only after 10 epochs on the validation set. 

\subsection{Results}
Table 1 shows the results from our experiment on the validation set of emotion classification. By using the pretrained weight from the EmotionNet \cite{wei2020learning} dataset, the model when only trains with seven emotion label gets the performance metric of 0.462. After apply the multi-task learning technique and using the pretrained weight from the VGGFace2 dataset, the metric improves by nearly 0.1. We also implement the shared backbone architecture between the two models of two tasks and get the performance metric of 0.713. The Focal loss which is used to counter the effect of imbalance dataset of seven emotions achieved the best results of 0.757. 
Table 2 displays the results from our experiments on the validation set of action unit detection. The application of shared backbone architecture does not improve the performance from using only annotation from the twelve action units in this case. However, the multi-task learning technique helps us attain the performance metric of 0.731.

\begin{table*}[ht]
\centering
\caption{Seven basic emotion classification result on the validation set}
\label{table1}
\begin{tabular}{@{}lccc@{}}
\toprule
\multicolumn{1}{c}{\textbf{Expression}} & \textbf{F1\_Score} & \textbf{Accuracy} & \textbf{0.67* F1 + 0.33* Acc} \\ \midrule
Baseline                                         & 0.3   & 0.5   & 0.366 \\
ResNet50 (EmotionNet)                                & 0.395 & 0.598 & 0.462 \\
ResNet50 (VGG-Face2) (Multitasking)              & 0.494 & 0.684 & 0.556 \\
ResNet50 (VGG-Face2) (Shared backbone)           & 0.675 & 0.791 & 0.713 \\
ResNet50 (VGG-Face2) (Multitasking) (Focal Loss) & 0.724 & 0.826 & 0.757 \\ \bottomrule
\end{tabular}
\end{table*}

% \usepackage{booktabs}
\begin{table*}[ht]
\centering
\caption{Facial action unit detection result on the validation dataset}
\label{tab2}
\begin{tabular}{@{}lccc@{}}
\toprule
\multicolumn{1}{c}{\textbf{Action Units}} & \textbf{F1\_Score} & \textbf{Accuracy} & \textbf{0.5* F1 + 0.5* Acc} \\ \midrule
Baseline                               & 0.22  & 0.4   & 0.31  \\
ResNet50 (EmotionNet)                      & 0.439 & 0.878 & 0.659 \\
ResNet50 (VGG-Face2) (Shared backbone) & 0.427 & 0.883  & 0.655 \\
ResNet50 (VGG-Face2) (Multitasking)    & 0.566 & 0.895 & 0.731 \\ \bottomrule
\end{tabular}
\end{table*}

\section{Conclusion}
In this paper, we present our experiments on the two tasks of emotion classification and action unit detection. ResNet50 with pretrained weight on the VGGFace2 dataset produces good results on the AffWild dataset and proposed training scheme with the application of multi-task learning enhances the performance by a considerable margin on both tasks. Moreover, Focal loss is suitable for solving the imbalance problem on the dataset of seven emotions.

\bibliography{ref.bib}{}
\bibliographystyle{unsrt}
\end{document}